\documentclass{article}

\usepackage[utf8]{inputenc}
\usepackage[T1]{fontenc}
\usepackage{lmodern}
\usepackage{graphicx}
\usepackage{booktabs}
\usepackage{array}
\usepackage{makecell}
\usepackage{amsmath}
\usepackage{float}
\usepackage{url}
\date{}
\usepackage{geometry}
\usepackage{subcaption}
\usepackage{appendix}

\title{Learning Reactive Human Motion Generation from Paired Interaction Data Using Transformer-Based Models}
\author{
\begin{tabular}{cc}
Masato Soga & Ryuki Takebayashi \\
Faculty of Systems Engineering & Graduate School of Systems Engineering \\
Wakayama University & Wakayama University
\end{tabular}
}

\begin{document}

\maketitle

\begin{abstract}
Recent advances in deep learning have enabled the generation of videos from textual descriptions as well as the prediction of future sequences from input videos. Similarly, in human motion modeling, motions can be generated from text or predicted from a single person's motion sequence. However, these approaches primarily focus on single-agent motion generation.
In contrast, this study addresses the problem of generating the motion of one person based on the motion of another in interaction scenarios, where the two motions are mutually dependent. We construct a dataset of paired action--reaction motion sequences extracted from boxing match videos and investigate the effectiveness of Transformer-based models for this task.
Specifically, we implement and compare three models: a simple Transformer, iTransformer, and Crossformer. In addition, we introduce a person ID embedding to explicitly distinguish between individuals, enabling the model to maintain structural consistency and better capture interaction dynamics.
Experimental results show that the simple Transformer can generate plausible interaction-aware motions without suffering from posture collapse, while iTransformer and Crossformer accumulate errors over time, leading to unstable motion generation. Furthermore, the proposed person ID embedding contributes to preventing structural collapse and improving motion consistency. These results highlight the importance of explicitly modeling individual identity in interaction-aware motion generation.
\end{abstract}

\section{Introduction}

Recent advances in deep learning have led to significant progress in human motion generation. Representative approaches include methods that generate motions from textual descriptions and methods that predict future motions from time-series motion data. In particular, research on motion prediction has been actively studied in recent years.

Martinez et al. revisited existing RNN-based approaches for human motion prediction and proposed a simple LSTM model incorporating residual connections \cite{martinez2017human}. Their results demonstrated that the proposed model achieves high accuracy in short-term prediction, even when compared with more complex conventional models. However, LSTM-based models rely on recurrent structures that require sequential processing, making parallel computation difficult. Moreover, while LSTM is effective at modeling short-term dependencies, it has limitations in capturing long-range dependencies.

In contrast, the Transformer \cite{vaswani2017attention}, based on attention mechanisms, can directly model relationships between distant elements and is well-suited for learning long-range dependencies. Owing to these advantages, various Transformer-based approaches for time-series modeling have been proposed. For example, Aksan et al. introduced a Transformer-based model for 3D human pose prediction, known as the Spatio-Temporal Transformer \cite{aksan2021spatiotemporal}. Unlike conventional autoregressive approaches, which suffer from error accumulation over time, their method explicitly models temporal and spatial dependencies using attention mechanisms, enabling stable long-term motion prediction.

Although these studies have demonstrated the effectiveness of Transformer-based models for human motion prediction, most existing approaches focus on single-person motion. Recently, motion generation conditioned on textual input has also been extensively studied; however, these approaches similarly focus on generating motions for a single individual.

In contrast, research on generating interactive motions between two individuals---where the motions are related but different---remains limited. In particular, the problem of generating a reaction sequence of one person given the action sequence of another person has not been sufficiently explored. Such models have practical applications in interactive domains such as video games, where real-time response generation is required, as well as in training systems for pair-based activities such as social dancing, where users can practice alone by observing the motion of a virtual partner.

Among the limited number of studies addressing this problem, Interaction Transformer\cite{chopin2023interaction} has been proposed as a Transformer-based model that generates a reaction sequence corresponding to an input action sequence given the initial pose of the reacting person. While Interaction Transformer has demonstrated the feasibility of generating reactive motions using datasets such as SBU~\cite{yun2012sbu}, the limited diversity of such datasets makes the model prone to overfitting. Furthermore, interaction patterns in which the distance between two individuals remain relatively unchanged, such as defensive ducking in response to punching, have not been sufficiently evaluated.

In addition, diffusion-based approaches such as ReMoS\cite{ghosh2023remos} have been proposed for generating reaction sequences that correspond to given action sequences, particularly in domains such as dance, including fine-grained motions of the fingers. While these approaches can generate diverse and highly realistic motions, their high computational cost makes real-time applications difficult.

To address these limitations, this study focuses on boxing as a domain that requires rapid generation of reaction motions. A dataset is constructed from boxing match videos obtained from Kaggle~\cite{kaggle_boxing}. The objective of this study is to investigate the capability of a simple Transformer, defined as a standard Transformer architecture directly applied to time-series motion generation, to generate appropriate reaction sequences for four types of boxing action sequences. Furthermore, to examine the effectiveness of different Transformer-based architectures, comparative experiments are conducted using iTransformer and Crossformer.

The main contributions of this study are summarized as follows:
\begin{itemize}
\item We formulate the task of generating reactive motion sequences of a counterpart given the action sequence of a subject in human-human interaction scenarios, focusing on boxing as a representative domain.
\item We construct a dataset of paired action-reaction motion sequences from publicly available boxing videos, and establish a preprocessing pipeline for extracting structured motion data suitable for learning.
\item We investigate the applicability of Transformer-based architectures to the proposed task by employing a simple Transformer model for time-series motion generation.
\item We introduce a person ID embedding to explicitly distinguish between individuals, improving structural consistency and stabilizing motion generation in interaction scenarios.
\item We conduct a comparative study using iTransformer and Crossformer, which have not previously been applied to interactive motion generation, and analyze their characteristics in this context.
\item We perform extensive experiments, including evaluations under both real-time and offline settings, to analyze factors affecting motion continuity, realism, and interaction consistency.
\end{itemize}
This study provides new insights into the role of identity-aware modeling in stabilizing interaction-based motion generation.

\section{Related Work}
In this section, we review related work by categorizing existing approaches according to their input modalities and problem settings, namely reaction motion generation, text-to-motion generation, and motion-to-motion or action-conditioned generation. This categorization clarifies the differences between existing methods and highlights the position of our approach.
\subsection{Reaction Motion Generation}
Studies that generate a partner's motion based on a given individual's motion are still limited in number, but several works have recently emerged.
\subsubsection{Interaction Transformer}
As a Transformer-based model that does not employ diffusion models, Interaction Transformer for Human Reaction Generation (hereafter referred to as InterFormer) \cite{chopin2023interaction} generates a corresponding reaction sequence by taking an action sequence of one person and the initial pose of the counterpart as inputs during inference.

\begin{figure}[t]
  \centering
  \includegraphics[width=0.8\linewidth]{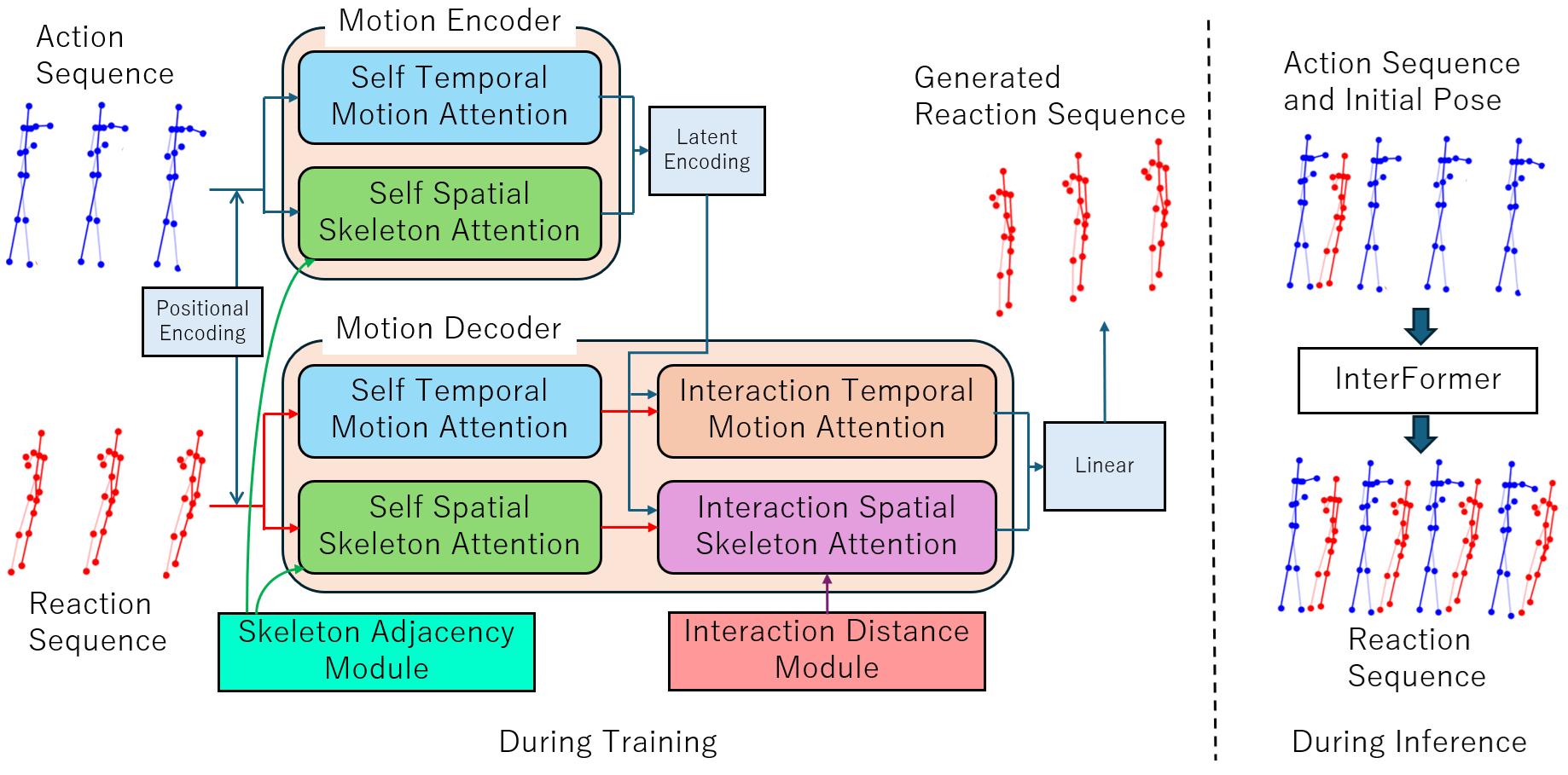}
   \caption{Overview of the InterFormer architecture during training (left) and inference (right) (adapted from \cite{chopin2023interaction})}
  \label{fig:InterFormer}
\end{figure}

The architecture of InterFormer, as well as its input/output configurations during training and inference, are illustrated in figure~\ref{fig:InterFormer}, adapted from the original paper. The encoder consists of two parallel self-attention mechanisms: Self Temporal Motion Attention, which is applied along the temporal axis, and Self Spatial Skeleton Attention, which models relationships among body joints at the same time step. This design is considered to be derived from prior work on spatio-temporal Transformer\cite{aksan2021spatiotemporal}.

During training, only the action sequence of one person is input to the encoder, while the reaction sequence is not provided to the encoder. In contrast, the decoder also consists of parallel modules of Self Temporal Motion Attention, which operates along the temporal dimension, and Self Spatial Skeleton Attention, which captures relationships among body joints at the same time step. During training, only the counterpart's reaction sequence is input to these decoder modules. Furthermore, the decoder includes Interaction Temporal Motion Attention and Interaction Distance Skeleton Attention. The latent representation of the action sequence, obtained after passing through the encoder and a latent encoding stage, is fed into these interaction attention modules. In addition, the vector representation of the reaction sequence, processed through the decoder's Self Temporal Motion Attention and Self Spatial Skeleton Attention, is also input to the interaction attention modules. Through this mechanism, the model learns both the temporal relationships between the action and reaction sequences and the spatial relationships based on inter-joint distances.

A notable feature of InterFormer is the introduction of the Interaction Distance Module, which explicitly computes the spatial distance between two persons. Specifically, it calculates pairwise distances between joints of the two skeletons and feeds the minimum joint distance into the decoder's Interaction Spatial Skeleton Attention. This design reflects the intuition that interactions become more active when two individuals are closer and diminish as they move apart. The Interaction Distance Module is therefore considered to facilitate efficient learning of such interaction dynamics.

Additionally, a Skeleton Adjacency Module is employed to provide information about joint connectivity to both the encoder's and decoder's Self Spatial Skeleton Attention modules. Since InterFormer feeds the action sequence only into the encoder and the reaction sequence only into the decoder, it is necessary to supply structural information separately to both. Given that the distances between connected joints in the human body remain constant, this module likely helps prevent the accumulation of joint position errors and the collapse of skeletal structure during motion generation.

InterFormer was evaluated using three datasets: SBU~\cite{yun2012sbu}, DuetDance~\cite{kundu2020crossconditioned}, and K3HI~\cite{xia2012k3hi}. In the punching scenario of the SBU dataset, when a punching action is given as the input action sequence, the generated reaction sequence exhibits a backward recoiling motion, outperforming four baseline models. This behavior is likely due to the fact that the SBU dataset contains only scenes where the counterpart moves backward in response to punching. Therefore, the Interaction Distance Module is considered to have played an important role, as the distance between two individuals is a crucial parameter in learning such retreating motions.

InterFormer was trained on a high-performance computing environment: Torch 1.8.1 on a PC with two 2.3 GHz processors, 64 GB RAM, and an NVIDIA Quadro RTX 6000 GPU. This computational setup likely enabled smooth motion generation despite the complexity of the model.

\subsubsection{ReMoS}

Next, we review ReMoS (Reactive Motion Synthesis) as a diffusion-based approach\cite{ghosh2023remos}. This model focuses primarily on dance as its domain and generates reaction sequences corresponding to action sequences, including detailed hand and finger movements.

ReMoS employs a DDPM (Denoising Diffusion Probabilistic Model)~\cite{ho2020ddpm} framework for both training and generation. In addition, the authors constructed a new dataset, ReMoCap, specifically for this study. This dataset contains approximately 275K motion frames and exceeds two hours of motion data. For comparison, the widely used SBU dataset contains only about 7K motion frames (approximately 0.13 hours), making it significantly smaller.

The model was trained on a system equipped with an NVIDIA RTX A4000 GPU, and the training process required approximately 8 to 11 hours. Due to the use of diffusion models, ReMoS achieves higher motion diversity compared to non-diffusion approaches. As stated in the original paper:``We observe that DDPMs such as ReMoS and ComMDM~\cite{shafir2023humanmotiondiffusion} have higher motion diversity compared to transformer-based methods such as InterFormer.''
However, the use of DDPM also results in longer generation times. For the ReMoCap dataset, generation can take up to 50 seconds, while for the 2C~\cite{shen2020interaction} and ExPI datasets~\cite{guo2022expi}, it can take up to 25 seconds.

\subsection{Text-to-Motion Generation}
A variety of models have been proposed for text-to-motion generation. In this section, we focus on three representative approaches.
\subsubsection{MotionDiffuse}
MotionDiffuse~\cite{zhang2022motiondiffuse} is one of the first deep learning models that adopts a denoising diffusion probabilistic model (DDPM)~\cite{ho2020ddpm} to generate 3D human motions conditioned on textual descriptions. By leveraging diffusion models, it achieves superior motion diversity and quality compared to earlier approaches based on generative adversarial networks (GANs). While MotionDiffuse does not explicitly report inference time, diffusion-based methods are generally known to require longer generation times due to iterative sampling procedures. Furthermore, MotionDiffuse enables fine-grained motion synthesis by modeling body-part-level dynamics, allowing the generation of composite motions (e.g., `a person is drinking water while walking'). However, it focuses on intra-person motion composition and does not address interactions between multiple individuals. This limitation motivates our approach.

\subsubsection{T2M-GPT}
T2M-GPT~\cite{zhang2023t2mgpt} generates human motion by first discretizing continuous motion sequences into discrete tokens using VQ-VAE, and then modeling these token sequences with a GPT-based architecture. Although this framework combines established techniques, it achieves higher-quality motion generation compared to diffusion-based approaches such as MotionDiffuse. By reformulating motion generation as a language modeling problem, T2M-GPT enables effective sequence modeling of human motion. However, it focuses on single-person motion generation conditioned on text and does not consider interactions between multiple individuals. This limitation motivates our approach.

\subsubsection{MoMask}
MoMask~\cite{guo2024momask} first converts motion into multi-level discrete tokens using Residual VQ. It then employs a masked transformer to generate coarse motion tokens by predicting randomly masked tokens conditioned on text. Subsequently, a residual transformer refines the motion representation by predicting higher-level residual tokens conditioned on both text and lower-level tokens.This coarse-to-fine generation process follows a progressive refinement paradigm, which is conceptually similar to diffusion-based methods. However, unlike diffusion models that refine continuous representations through stochastic denoising, MoMask performs deterministic refinement in a discrete token space using residual prediction.

MoMask is evaluated on standard benchmarks such as HumanML3D~\cite{guo2022humanml3d} and KIT-ML~\cite{plappert2016kitml}, where it outperforms prior methods such as MotionDiffuse and T2M-GPT in terms of text-motion alignment. However, its motion diversity is relatively lower compared to diffusion-based approaches. In addition, MoMask requires a fixed-length token sequence, which limits its flexibility in generating variable-length motions. This limitation may restrict its applicability to real-world scenarios where motion durations vary.

Unlike these methods, our approach conditions on another person's motion rather than textual descriptions.

\subsection{Motion-to-Motion and Action-conditioned Generation }
While motion prediction models take previous motion as input, action-conditioned methods rely on high-level action labels. However, both approaches focus on single-person motion generation and do not explicitly model interactions between multiple individuals.

\subsubsection{Spatio-Temporal Transformer}
Single-person motion generation has been widely studied. A representative approach is the spatio-temporal Transformer~\cite{aksan2021spatiotemporal}, which takes a sequence of past motion frames as input and predicts future motion frames by modeling both temporal dependencies and spatial relationships among joints. These approaches have demonstrated strong performance in motion prediction tasks; however, they do not explicitly consider interactions between multiple individuals.
While the standard Transformer~\cite{vaswani2017attention} applies self-attention along the temporal dimension, spatio-temporal transformers introduce two parallel attention mechanisms: temporal attention, which captures dependencies over time, and spatial attention, which models relationships among joints at each time step. By applying both attention mechanisms to motion features with temporal positional encoding, the model can effectively learn both temporal and spatial dependencies in human motion. This design has been widely adopted in subsequent motion generation models, including InterFormer.

Although full spatio-temporal attention (i.e., attention over all pairs of joints across all time steps) considers all pairwise relationships, it suffers from increased computational complexity and a lack of structural inductive bias. In contrast, spatio-temporal attention decomposes the problem into temporal and spatial components, enabling more effective and stable learning.

\subsubsection{ACTOR}
Action-conditioned motion generation methods, such as ACTOR~\cite{petrovich2021actor}, generate human motion from discrete action labels representing predefined motion categories. Unlike text-to-motion approaches that rely on natural language descriptions, these methods use structured categorical inputs, resulting in more constrained but controllable motion generation.
ACTOR adopts a Transformer-based variational autoencoder (VAE) framework to model human motion. During training, the encoder takes both the motion sequence and the corresponding action label to learn a latent representation of motion. During inference, the model generates motion conditioned only on the action label by sampling from the learned latent space.
Although ACTOR enables controllable motion generation from action labels, it focuses on single-person motion and does not explicitly model interactions between multiple individuals. This limitation motivates our approach, which explicitly models interaction between multiple individuals.

\subsubsection{TransFusion}
TransFusion~\cite{zhang2023transfusion} is a diffusion-based motion generation model that learns to generate motion sequences by iteratively denoising random noise. Unlike deterministic motion prediction models that generate a single future sequence conditioned on past motion, TransFusion adopts a stochastic generation process and can incorporate conditioning information such as an input motion sequence.
During inference, given an input motion sequence, the model generates multiple plausible future motion sequences that are temporally consistent with the input, rather than a single deterministic continuation. Compared to earlier approaches, TransFusion achieves a better balance between motion accuracy and diversity.

\section{Employed Model}
We employed three models: a simple Transformer, iTransformer, and Crossformer. Here, the simple Transformer refers to a standard Transformer architecture applied directly to time-series motion generation. The iTransformer and Crossformer are improved Transformers for specific purposes. 

\subsection{iTransformer}
Yong Liu et al. proposed iTransformer \cite{liu2023itransformer}, a Transformer-based model for multivariate time-series data. Unlike conventional Transformers, which apply attention along the temporal dimension, iTransformer is characterized by applying attention across the feature dimension of the input. This design enables the model to effectively capture dependencies among features and has demonstrated strong performance in multivariate time-series forecasting.
In this study, we hypothesize that such capability to model inter-feature dependencies is beneficial for learning interactions between the motions of two individuals, as well as the relationships among joints in human pose data. Based on this idea, we construct a model based on iTransformer and use it as a comparative baseline against the simple Transformer.

\subsection{Crossformer}
Similar to iTransformer, Zhang et al. proposed Crossformer~\cite{zhang2023crossformer}, a Transformer-based model for multivariate time-series data. To capture long-term dependencies inherent in time-series data, Crossformer introduces a learning framework based on a Two-Stage Attention (TSA) layer.
In this study, we consider that the ability of Crossformer to jointly model dependencies along both the temporal dimension and the feature dimension is effective for learning interactions between the motions of two individuals, as well as inter-joint and temporal relationships in human pose data. Based on this perspective, we construct a model based on Crossformer and use it as a comparative baseline against the simple Transformer. Section~\ref{sec:ProposedMethod} provides a detailed description of the architectures of these Transformer-based time-series models.

\section{Proposed Method}
\label{sec:ProposedMethod}
\subsection{Overview of the Proposed Method}
In this study, we construct three Transformer-based models that generate the motion of a counterpart corresponding to a subject's own motion. Prior to training, we newly create a dataset suitable for this task. Specifically, we estimate the poses of two individuals engaged in a boxing match from video data and generate 3D pose data in which the three-dimensional coordinates of each joint are used as features. Furthermore, preprocessing and data formatting are applied to the generated 3D pose data. The models are then trained using the constructed dataset.
In addition, we employ the trained models together with motion capture and Unity to perform real-time motion generation and visualization. Through these experiments, we evaluate the effectiveness of the proposed method and discuss the results.

\subsection{Development Environment}

The development environment is as follows. The PC used in this study was a Lenovo Legion T7 34IMZ5 equipped with an Intel Core i9-14900HX, an NVIDIA GeForce RTX 4060, 32.0 GB of memory, and Windows 11 Home. The same PC was used throughout the entire process, from development to evaluation experiments. Python version 3.10.0 was used for software development, and PyTorch version 2.3.1+cu118 was employed for constructing the deep learning models.

To capture user motion, Azure Kinect DK \cite{azurekinect} was used. As the SDK for processing depth and IR images obtained from Azure Kinect DK and providing body tracking results, Azure Kinect Body Tracking SDK version 1.1.2 was utilized.  Among the body joint positions obtainable from the Azure Kinect Body Tracking SDK, this study uses 27 joints excluding five points (left and right eyes, ears, and nose), resulting in a total of 81 three-dimensional skeletal coordinates as input features. These five points were excluded because they have little influence on body motion, with the aim of improving computational efficiency and prediction performance of the learning model. In addition, the Azure Kinect Body Tracking SDK was operated in GPU mode in this study.  It should be noted that the software developed using Azure Kinect DK was used only for the real-time evaluation described in Section~\ref{sec:RealTimeEvaluation}. It was not used for collecting training data for the deep learning model. The training data were obtained from boxing videos, as described in Section~\ref{sec:VideoDataset}.

Furthermore, Unity version 2019.1.2f1 was used to develop the interface for visualizing motion, and the scripts were written in \texttt{C\#}. In addition, stream-based socket communication based on TCP (Transmission Control Protocol) was adopted for data transmission between the deep learning model and Unity. This visualization interface was used both for displaying the user's motion during the real-time evaluation in Section~\ref{sec:RealTimeEvaluation} and for displaying the counterpart motion generated by the deep learning model in both the real-time evaluation and the offline evaluation described in Section~\ref{sec:OfflineEvaluation}.

\subsection{Training data}
The training data used in this study consist of 3D pose data generated from boxing match videos. First, 2D pose estimation is performed on the video dataset to extract the poses of the two individuals engaged in the match. Next, 3D pose estimation is conducted based on the extracted 2D pose data. Furthermore, preprocessing and data formatting are applied to the obtained 3D pose data to construct input data for training the deep learning model. This section describes the method for generating these training data.

\subsubsection{Video Dataset}
\label{sec:VideoDataset}
In this study, we use the Olympic Boxing Video Dataset \cite{kaggle_boxing}, which is publicly available on Kaggle. This dataset consists of video recordings of boxing matches and includes 21 match videos. The videos have a frame rate of 50 fps and a resolution of 1920 $\times$ 1080 pixels.  Pose estimation is performed on these videos on a frame-by-frame basis to generate the training data.

\subsubsection{Extraction of Match Segments}
The video dataset described in the previous section contains segments in which boxing matches are not taking place. Including such segments in the training data would degrade the quality of the dataset. Therefore, in this study, we perform video cutting to segment the videos and extract only the match intervals.

The segments in which matches are not taking place can be broadly classified into four types. The first type corresponds to the interval between the end of one match and the start of the next match. The second type is the interval between rounds, as each match consists of three rounds. The third type refers to judging segments in which decisions or confirmations are made by the judges. The fourth type consists of periods in which the match is temporarily suspended by the referee.  In this study, these segments are excluded, and only the intervals in which the match is actually in progress are used for pose estimation.

\subsubsection{2D Pose Estimation}
For 2D pose estimation, AlphaPose \cite{fang2020alphapose} was used as the framework. AlphaPose is a multi-person 2D pose estimation framework that integrates human detection and 2D pose estimation for each detected person. It is designed to achieve high accuracy by incorporating existing 2D pose estimators.

\begin{figure}[t]
 \centering
  \includegraphics[width=0.4\linewidth]{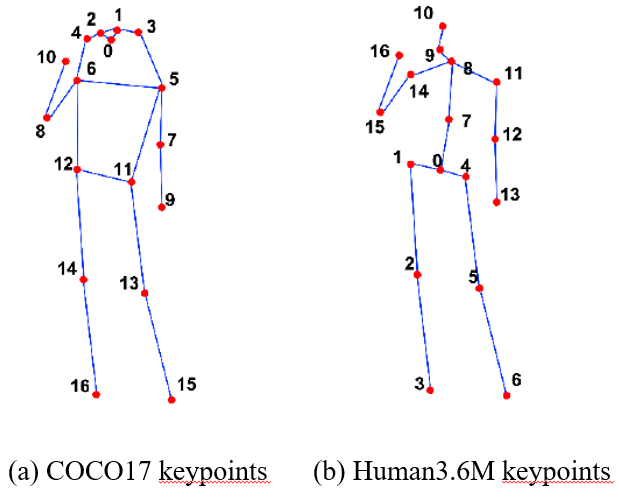}
  \caption{Keypoint structures of COCO17 and Human3.6M}
  \label{fig:coco17}
\end{figure}

In this study, FastPose was selected as the internal 2D pose estimator, and ResNet-152 was used as its backbone (feature extractor). The output consists of 2D coordinates based on the COCO-17 (17 joints) human body model definition. Figure~\ref{fig:coco17} (a) shows the configuration of the COCO-17 keypoints \cite{mmpose2d}.

YOLO-X was used as the human detector, and OC-SORT was employed as the tracker. However, the boxing video dataset used in this study contains not only the competitors but also spectators and referees, who frequently cross the frame. As a result, frequent ID switches were observed. Therefore, to ensure temporal consistency for each individual, manual ID consolidation was performed as a post-processing step.

In addition, prior to 3D pose estimation, preprocessing was applied to the 2D pose time-series data. The deep learning model for motion generation used in this study requires fixed-length time-series data (30 frames) as input; therefore, the input data must consist of at least 30 consecutive frames. However, during the processes of human tracking and 2D pose estimation, frame drops caused by temporary detection failures or occlusions were observed. These missing frames not only degrade the accuracy of 3D pose estimation but also significantly reduce the amount of valid training data due to the fixed input length constraint.

To address this issue, in order to increase the amount of data and maintain temporal continuity, linear interpolation was applied using the joint coordinates of the preceding and succeeding frames, but only when the length of the missing segment was within three consecutive frames. This is based on the assumption that, for short-term missing intervals, joint motion can be approximately modeled as linear, and thus the negative impact on pose estimation accuracy is limited.

Furthermore, even after linear interpolation, time-series data with fewer than 30 consecutive frames were excluded from the training dataset.  The linear interpolation is defined by Equation (1), where $p[t]$ represents a vector whose components are the 2D coordinates of each joint at time $t$, and $\alpha$ is a coefficient representing the relative position of the missing frame $t$ between the preceding and succeeding frames $t_0$ and $t_1$.

\begin{equation}
p[t] = (1 - \alpha)\, p(t_0) + \alpha\, p(t_1), \quad
\alpha = \frac{t - t_0}{t_1 - t_0}
\end{equation}
\subsubsection{3D Pose Estimation}
For 3D pose estimation, MotionBERT \cite{zhu2023motionbert} was employed. MotionBERT is a pose lifting method that takes as input a sequence of 2D joint coordinates across multiple frames and estimates the 3D joint coordinates for each frame.

In this study, the sequence of 2D joint coordinates based on COCO-17, obtained in the previous section, was used as input. However, since MotionBERT is trained based on the Human3.6M human body model definition, joint mapping was performed according to the correspondence between COCO-17 and Human3.6M. Table~\ref{tab:JointMapping} presents the mapping between the Human3.6M joint vector $H[i]$ and the COCO-17 joint vector $C[i]$. Figure~\ref{fig:coco17} (b) shows the configuration of the Human3.6M keypoints \cite{mmpose3d}.

\begin{table}[t]
  \centering
  \caption{Joint mapping definition between Human3.6M and COCO-17}
  \label{tab:JointMapping}
  \begin{tabular}{c|c|p{7cm}}
    \toprule
    Human3.6M & COCO-17 & Notes\\
    \midrule
    $H[0]$ & $(C[11]+C[12])/2$ & The pelvis position of Human3.6M is computed as the average of the left and right hip joint positions of COCO-17.\\
    \midrule
    $H[1]$ & $C[12]$ & Right hip joint\\
    $H[2]$ & $C[14]$ & Right knee\\
    $H[3]$ & $C[16]$ & Right ankle\\
    $H[4]$ & $C[11]$ & Left hip joint\\ 
    $H[5]$ & $C[13]$ & Left knee\\
    $H[6]$ & $C[15]$ & Left ankle\\
    \midrule
    $H[7]$ & \makecell{$Cpelvis=(C[11]+C[12])/2$\\   $Cneck=(C[5]+C[6])/2$\\   $Ctrunk\_center=(Cpelvis+Cneck)/2$} & The trunk center of Human3.6M is computed as the average of the pelvis and neck positions of COCO-17.\\
    \midrule
    $H[8]$ & $(C[5]+C[6])/2$ & The neck position is computed as the average of the left and right shoulder positions of COCO-17.\\
    \midrule
    $H[9]$ & $C[0]$ & Nose\\
    \midrule
    $H[10]$ & \makecell{$Cneck=(C[5]+C[6])/2$\\ $Cheadtop=C[0]+(C[5]+Cneck)$} & The head top position of Human3.6M is approximated by extending the direction vector from the nose to the neck by the same length from the nose position of COCO-17.\\
    \midrule
    $H[11]$ & $C[5]$ & Left shoulder\\
    $H[12]$ & $C[7]$ & Left elbow\\
    $H[13]$ & $C[9]$ & Left wrist\\
    $H[14]$ & $C[6]$ & Right shoulder\\ 
    $H[15]$ & $C[8]$ & Right elbow\\
    $H[16]$ & $C[10]$ & Right wrist\\        
    \bottomrule
  \end{tabular}
\end{table}

\begin{figure}[b]
  \centering
   \includegraphics[width=0.2\linewidth]{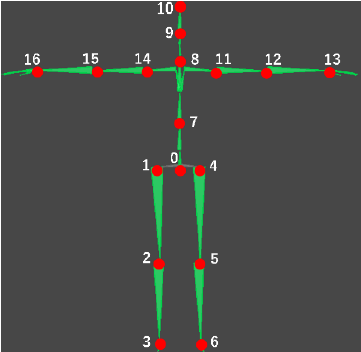}
  \caption{Keypoint structure of retargeted Humanoid model}
  \label{fig:Humanoid}
\end{figure}

\subsubsection{Preprocessing of Training Data}
\label{sec:PreprocessingOfTrainingData}

In this study, preprocessing was applied to the 3D pose data represented in the Human3.6M format to obtain a representation suitable for learning.

First, to reduce the influence of differences in body proportions, retargeting to a predefined body configuration was performed on the 3D pose data. Specifically, the joints of a pre-prepared Humanoid model were mapped to the 17 joints of the Human3.6M representation, and the bone lengths between joints in the Humanoid model were obtained. Then, for the 3D joint coordinates in the Human3.6M format used as training data, each joint was translated such that the distance between connected joints matches the bone lengths of the target Humanoid model, while preserving the original direction between joints. Figure~\ref{fig:Humanoid} shows the configuration of the joints of the Humanoid model used in this study. In addition, the joint vector of Human3.6M, $H[i]$, and that of the target Humanoid model, $R[i]$, are in one-to-one correspondence based on the same joint indices, and in this study, all joints are treated as satisfying 
$H[i]=R[i]$.

Next, normalization was performed for all frames using the pelvis position ($H[0]$) as the reference. Specifically, the pelvis position in each frame was translated to $(x,y,z)=(0,1,0)$, and the same offset was applied to all joints within the same frame. As a result, the 3D pose is represented as a relative pose with respect to the pelvis position.

Finally, to suppress the effect of numerical errors and to unify the representation precision of the data, all 3D joint coordinates were rounded to six significant digits. Through the above preprocessing steps, the 3D pose data are transformed into a standardized representation that eliminates the effects of body proportions and global position, making them suitable for use as input to the learning model constructed in this study. A total of 210,926 frames were ultimately obtained as the training dataset through this process.

\subsection{Learning Models}
In this study, three types of Transformer-based time-series learning models were employed to train a model that generates the motion of one individual based on the motions of two individuals as input. Specifically, we used a simple Transformer specialized for learning temporal dependencies, iTransformer specialized for learning dependencies among features, and Crossformer, which can simultaneously learn dependencies along both the temporal and feature dimensions. The architecture and training method of each model are described in detail below.

\subsubsection{Simple Transformer}
\begin{figure}[H]
  \centering
   \includegraphics[width=0.6\linewidth]{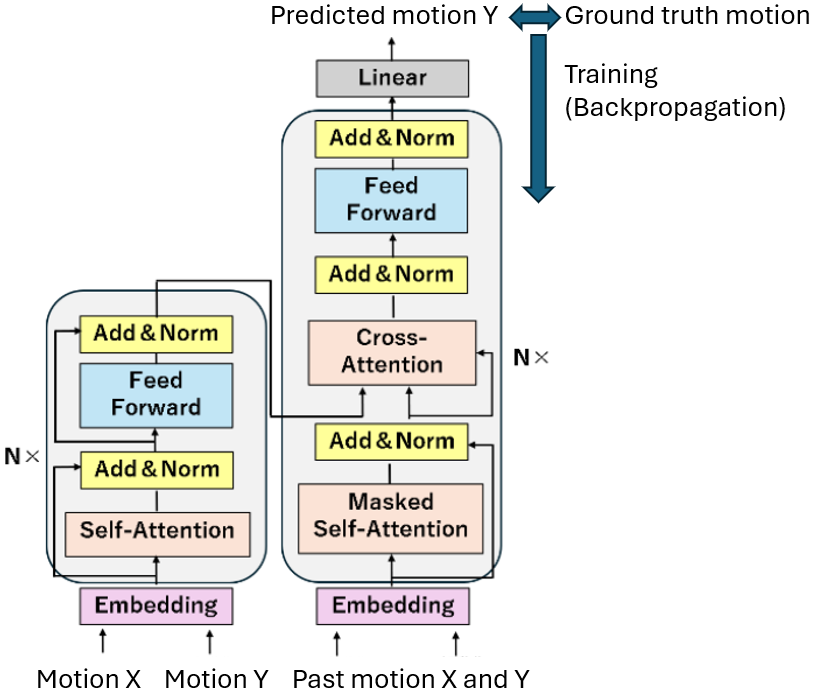}
  \caption{Motion prediction by simple Transformer}
  \label{fig:simpleTransformer}
\end{figure}

This model is a standard Transformer-based time-series learning model primarily designed to learn temporal dependencies in time-series data. The architecture of the model is shown in Figure~\ref{fig:simpleTransformer} .

In the encoder, the input motion sequences $X$ and $Y$, which are constructed from 3D pose data, are provided as inputs. Here, $X$ represents the subject's motion, and $Y$ represents the counterpart's motion. Each of $X$ and $Y$ is a vector representing joint features at each frame. When fed into the encoder, these inputs are concatenated along the feature dimension to form a single input sequence. The encoder then learns a feature representation of the entire time series based on this combined input.

The decoder takes as input the past motions of $X$ and $Y$, as well as their ground-truth motion. By referring to both the encoded feature representation and the past motion information, the decoder predicts future motion. The final output of the decoder corresponds to the predicted future motion.

Within the model, layers that apply self-attention along the temporal dimension are employed. These layers learn how strongly features at each time step are related to those at other time steps, thereby capturing temporal dependencies within the sequence. As a result, the model can obtain feature representations that incorporate long-term motion context.

The dimensionality of each input is 17 (number of joints) $\times$ 3 (3D coordinates), resulting in a 51-dimensional vector. The sequence lengths are 30 frames for the input motions $X$  and $Y$, 10 frames for the past motions, and 1 frame for the ground-truth (target) motion.

\subsubsection{iTransformer}
\begin{figure}[H]
  \centering
   \includegraphics[width=0.45\linewidth]{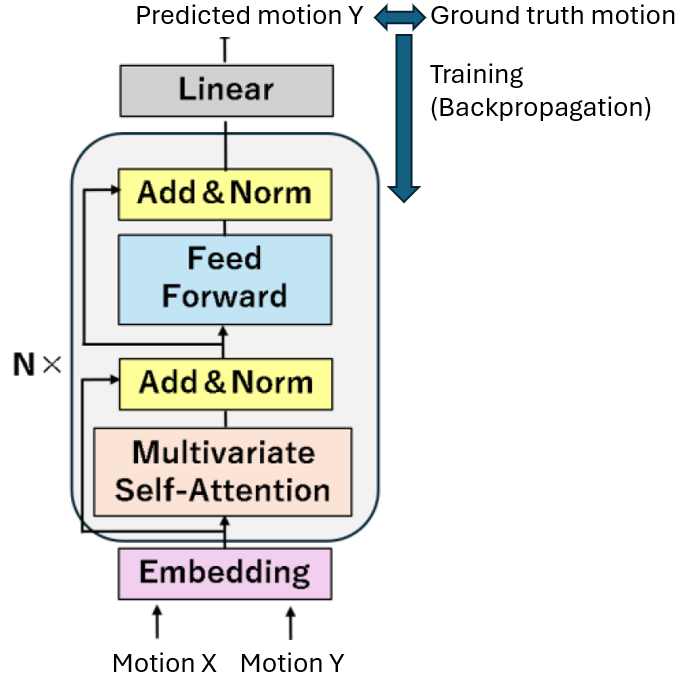}
  \caption{Motion prediction by iTransformer}
  \label{fig:iTransformer}
\end{figure}

ITransformer is a Transformer-based time-series learning model designed primarily to learn dependencies along the feature dimension in time-series data. The architecture of the model is shown in Figure~\ref{fig:iTransformer} .

The encoder takes as input the motion sequences $X$ and $Y$, which are constructed from 3D pose data. Unlike conventional Transformers, this model treats features (i.e., joints) as tokens rather than time steps. Specifically, during value embedding, the input motion sequences are restructured into feature vectors corresponding to each joint while preserving the temporal length. With this representation, each feature is treated as a token in the encoder, and self-attention is applied across the feature dimension. As a result, iTransformer captures dependencies in the feature space, such as joint configurations and inter-joint relationships, rather than temporal variations.

The feature representations obtained from the encoder are then passed through a linear projection to regress the 3D pose at future time steps. Concretely, based on the aggregated past time-series information encoded by the encoder, the 3D coordinates of each joint are directly predicted and output as the predicted motion sequence.

It should be noted that iTransformer is an encoder-only model and does not include a decoder. As described above, self-attention in iTransformer is applied along the feature dimension rather than the temporal dimension. Consequently, modeling of temporal dependencies is not handled by self-attention but is primarily delegated to the linear projection. In this process, past time-series information is aggregated along the temporal dimension through a weighted summation, where the contribution of each time step is determined by learned weights. As a result, temporal information necessary for prediction is integrated in a fixed manner. Depending on the characteristics of the time-series data, this often leads to relatively larger weights being assigned to temporally closer frames, resulting in a stronger influence of recent time steps.

The dimensionality of each input is 17 (number of joints) $\times$ 3 (3D coordinates), resulting in a 51-dimensional vector. The sequence lengths are 30 frames for the input motions $X$ and $Y$,, and 1 frame for the predicted motion.

\subsubsection{Crossformer}

\begin{figure}[t]
  \centering
   \includegraphics[width=0.8\linewidth]{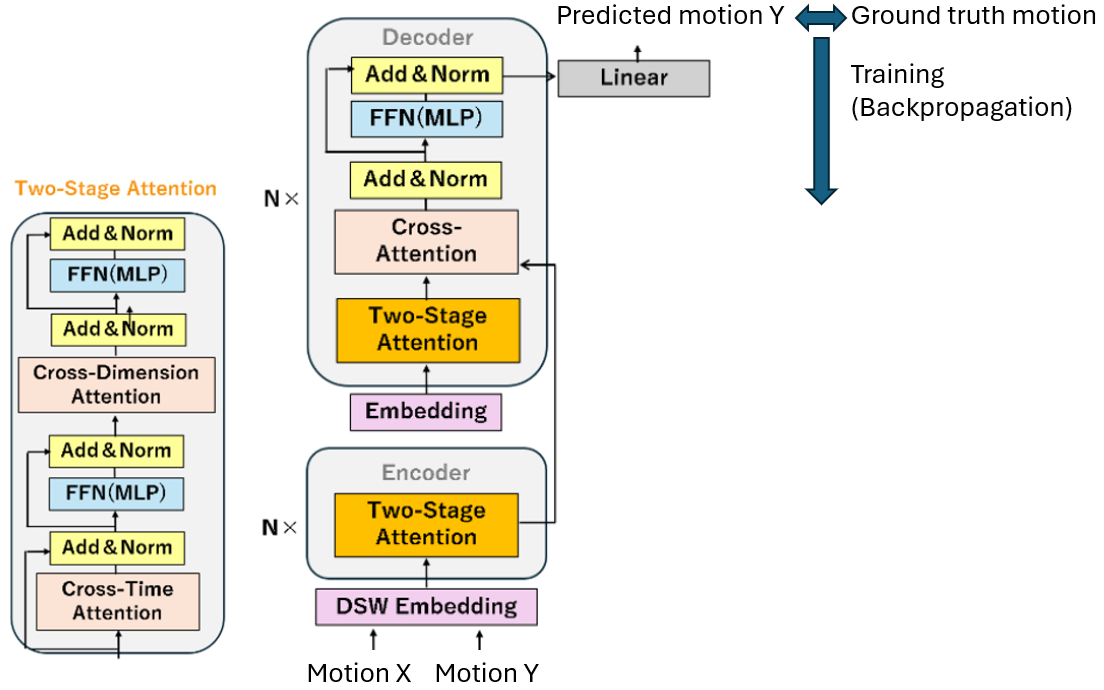}
  \caption{Motion prediction by Crossformer}
  \label{fig:crossformer}
\end{figure}

Crossformer is a Transformer-based time-series learning model designed to simultaneously learn dependencies along both the temporal and feature dimensions in time-series data. The architecture of the model is shown in Figure~\ref{fig:crossformer}.

The encoder takes as input the motion sequences $X$ and $Y$, which are constructed from 3D pose data. The input to the encoder is formed in the same manner as described in the simple Transformer section. However, during value embedding, the input is first transformed from time-series vectors into feature-wise vectors, similar to iTransformer. Then, for each dimension, the temporal axis is divided into fixed-length segments and vectorized using Dimension-Segment-Wise (DSW) embedding. 

The input to the decoder consists of learnable positional embeddings. Specifically, it is defined as a parameter tensor of shape $(B, D, out\_seg, d\_model)$, where 
$B$ is the batch size, 
$D$ is the number of features, 
$out\_seg$ is the number of segments obtained by dividing the prediction length by the segment length, and 
$d\_model$
 is the model dimensionality. The decoder generates future segments for each feature by referencing this embedding representation together with the encoder output.

Using these embedded representations, the encoder of Crossformer performs learning based on the Two-Stage Attention (TSA) mechanism.

In TSA, the Cross-Time Stage first applies self-attention across segments for each dimension. This enables the model to capture temporal motion patterns for each joint.

Next, in the Cross-Dimension Stage, a small number of learnable tokens (referred to as Routers) are introduced for each segment position to efficiently model inter-dimensional interactions. Specifically, a two-stage attention mechanism is employed, where the Routers aggregate information across dimensions as queries and then distribute the information back to each dimension. Through this design, the computational complexity of self-attention along the feature dimension is reduced from 
$O(D^2)$ to $O(FD)$, where 
$F$ denotes the number of Routers.

\subsubsection{Person ID Embedding}

In this study, to handle motion sequences of multiple individuals simultaneously and to improve prediction accuracy, we investigate a method for embedding person IDs. The purpose of person ID embedding is to explicitly provide the model with attribute information indicating which features belong to which individual when multiple persons' motions are mixed in the input.

Specifically, a learnable high-dimensional vector is assigned to each individual as a person ID embedding, and this vector is added to the entire input sequence. This encourages stronger interactions among features belonging to the same individual while suppressing interactions with features from different individuals, thereby enabling efficient learning of identity-specific temporal features.

While incorporating person ID embeddings after training progresses may enhance the learning of motion patterns specific to each individual, it may also weaken the learning of correspondence relationships between different individuals. Therefore, in the evaluation, we conduct experiments under two conditions: with and without person ID embedding.

\section{Evaluation}
\label{sec:Evaluation}

\subsection{Objective of the Evaluation}

The objective of this evaluation is to assess the quality of the generated motions. The evaluation focuses on three main aspects.

The first aspect is whether the motion is temporally continuous, i.e., whether the transitions between frames are smooth. The second aspect is whether the generated motion of the counterpart is a coherent and appropriate response to the subject's motion. The third aspect is an overall subjective evaluation of whether the motion appears human-like.

A questionnaire survey is conducted based on these criteria, and the results are analyzed and discussed accordingly.

\subsection{Overview of the Evaluation}

In the evaluation, each of the three models described in Section 3 is assessed. This section first describes the parameter settings used during model training, followed by the evaluation methodology.

\subsubsection{Model Parameters}

In this study, the basic training parameters were unified across all three models. Table~\ref{tab:params}  presents the training parameters of the models.

Note that the segment length and the number of routers, which represent the length of temporal segmentation and the number of learnable tokens in Crossformer, are configurable hyperparameters. The models were trained under multiple settings for these parameters, and the combination that was found to be effective in terms of performance was selected and used for the evaluation.

\begin{table}[H]
  \centering
  \caption{Model parameters}
  \label{tab:params}
  \begin{tabular}{cccc}
    \toprule
    Parameter & Value \\
    \midrule
    The number of encoders & 2 \\
    The number of decoders & 1\\
    The number of dimensions of the model & 512 \\
    The number of attention heads & 8 \\
    The number of dimensions of FFN & 2048 \\
    Ratio of dropout & 0.1\\
    The number of dimensions of the model & 512 \\
    Activation function & gelu \\
    Batch size & 32 \\
    Learning ratio & 0.0001 \\
    Lost function & MSE \\
    Segment length for only Crossformer & 5 \\
    The number of routers for only Crossformer & 30 \\
    \bottomrule
  \end{tabular}
\end{table}

\subsubsection{Motion Types}

In this evaluation, motions commonly observed in boxing were performed, and the quality of the generated motions was assessed. In this study, four representative input motions were defined and used for evaluation.

The first is an offensive motion: a straight punch directed toward the opponent (straight). The second is an offensive motion: a punch delivered from the side (hook). The third is a defensive motion: a guarding action in which the hands are held up in front of the face (block). The fourth is a defensive motion: an evasive action in which the upper body is lowered forward and downward (ducking).

These motions were selected because they frequently occur in boxing matches and involve significant changes in body posture, making them suitable for evaluation.

Furthermore, regarding the counterpart's motion generated in response to these inputs, it is assumed that defensive motions are generated for offensive inputs, while offensive motions are generated for defensive inputs.

\subsubsection{Real-Time Evaluation}
\label{sec:RealTimeEvaluation}

\begin{figure}[t]
  \centering
   \includegraphics[width=0.7\linewidth]{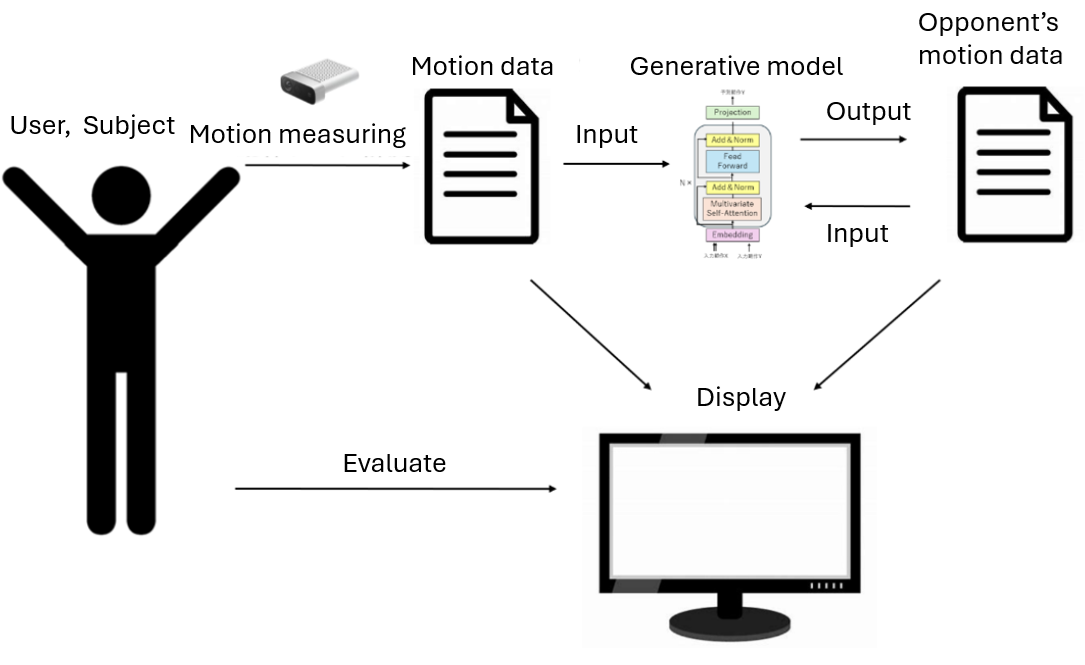}
  \caption{ System Pipeline}
  \label{fig:SystemPipeline}
\end{figure}

In this study, we first conduct an evaluation using a system that simultaneously displays the subject's motion and the generated counterpart motion in real time. Figure~\ref{fig:SystemPipeline} illustrates the system pipeline.

The system operates as follows: the user's motion is captured using Azure Kinect DK, and both the captured subject motion and the counterpart motion are input to the model in sequences of 30 frames to generate new counterpart motion. The captured subject motion and the generated counterpart motion are then sequentially visualized using humanoid avatars in Unity.  After the initial 30-frame input to the model, the subject motion input is continuously updated using the motion captured by Azure Kinect DK, while the counterpart motion input is updated using the motion generated by the model.

For the initial 30 frames, the counterpart motion is not yet generated by the model, and thus no input motion exists for the counterpart. Therefore, in this study, considering that boxing motions are typically performed with two individuals facing each other, the initial 30 frames of the subject's motion are mirrored along the x-axis and z-axis and used as the initial counterpart motion input to the model. Note that these initial 30 frames do not include offensive or defensive actions but represent a basic fighting stance in boxing. 
In addition, to ensure consistency between the body proportions of the training data and the motion data captured by the system, the captured motion data are preprocessed using the same method described in Section~\ref{sec:PreprocessingOfTrainingData} before being input to the model.

\subsubsection{Offline Evaluation}
\label{sec:OfflineEvaluation}

In the evaluation described in the previous section, the subject's motion is measured in real time. However, this approach may be affected by inference latency, which can degrade the accuracy of the generated motion. Therefore, in this section, we perform an offline evaluation using a portion of the dataset that was not used for training and reserved as test data.

Specifically, using data suitable for evaluation, both the subject's motion and the counterpart's motion for the initial 30 frames are obtained from the test data. After the initial 30-frame input to the model, the subject's motion input is sequentially updated using the test data beyond the initial frames, while the counterpart's motion input is updated using the motion generated by the model.

Finally, all frames of the subject's motion from the test data and all frames of the generated counterpart motion are visualized using humanoid avatars in Unity for evaluation.

\subsubsection{Questionnaire-Based Experiment}

In this study, a questionnaire survey was conducted as an experimental evaluation to assess the quality of the motions generated by the proposed models. As described in Sections 4.2.3 and 4.2.4, the survey consists of videos showing motions generated in real time and videos generated in an offline environment. Participants watch these videos and respond to the questionnaire. The videos presented to the participants consist of a single sequence in which the four input motions described in Section 4.2.2 are performed consecutively.

The same questionnaire items are used to evaluate each model described in Section 3. All responses are measured using a 5-point Likert scale, where 1 indicates ``strongly disagree'' and 5 indicates ``strongly agree.'' The questionnaire items are as follows:

(1) Are the transitions between frames of the generated counterpart motion temporally smooth?

(2) Is the generated counterpart motion a coherent and appropriate response to the subject's motion?

(3) Does the generated counterpart motion appear human-like overall?

\section{Results}

This section presents the results of the questionnaire survey described in Section~\ref{sec:Evaluation}, conducted with 17 participants.

\subsection{Results of the Real-Time Evaluation} 

Tables~\ref{tab:simpleTransformer_withoutID} to~\ref{tab:Crossformer_withID} present the questionnaire results for each model in the real-time evaluation. Table~\ref{tab:simpleTransformer_withoutID} shows the results for the simple Transformer (without person ID embedding), Table~\ref{tab:simpleTransformer_withID} for the simple Transformer (with person ID embedding), Table~\ref{tab:iTransformer_withoutID} for iTransformer (without person ID embedding), Table~\ref{tab:iTransformer_withID} for iTransformer (with person ID embedding), Table~\ref{tab:Crossformer_withoutID} for Crossformer (without person ID embedding), and Table~\ref{tab:Crossformer_withID} for Crossformer (with person ID embedding).

In addition, the corresponding statistical measures are presented in Tables~\ref{tab:SsimpleTransformer_withoutID} to~\ref{tab:SCrossformer_withID}.

\begin{table}[H]
  \centering
  \caption{Results of the Simple Transformer (without person ID embedding) in the real-time evaluation}
  \label{tab:simpleTransformer_withoutID}
  \begin{tabular}{|c|c|c|c|c|c|}
    \toprule
    Quest.No. & 1 strong disagree & 2 disagree & 3 Neutral & 4 agree & 5 strong agree \\
    \midrule
    $(1)$ & 0 & 2 & 3 & 6 & 6 \\
    $(2)$ & 1 & 4 & 7 & 4 & 1 \\
    $(3)$ & 0 & 5 & 4 & 7 & 1 \\
    \bottomrule
  \end{tabular}
\end{table}

\begin{table}[H]
  \centering
  \caption{Results of the Simple Transformer (with person ID embedding) in the real-time evaluation}
  \label{tab:simpleTransformer_withID}
  \begin{tabular}{|c|c|c|c|c|c|}

    \toprule
    Quest.No. & 1 strong disagree & 2 disagree & 3 Neutral & 4 agree & 5 strong agree \\
    \midrule
    $(1)$ & 1 & 0 & 1 & 11 & 4 \\
    $(2)$ & 0 & 4 & 7 & 4 & 2 \\
    $(3)$ & 0 & 0 & 4 & 11 & 2 \\
    \bottomrule

  \end{tabular}
\end{table}

\begin{table}[H]
  \centering
  \caption{Results of the iTransformer (without person ID embedding) in the real-time evaluation}
  \label{tab:iTransformer_withoutID}
  \begin{tabular}{|c|c|c|c|c|c|}

    \toprule
    Quest.No. & 1 strong disagree & 2 disagree & 3 Neutral & 4 agree & 5 strong agree \\
    \midrule
    $(1)$ & 4 & 1 & 5 & 3 & 4 \\
    $(2)$ & 17 & 0 & 0 & 0 & 0 \\
    $(3)$ & 17 & 0 & 0 & 0 & 0 \\
    \bottomrule

  \end{tabular}
\end{table}

\begin{table}[H]
  \centering
  \caption{Results of the iTransformer (with person ID embedding) in the real-time evaluation}
  \label{tab:iTransformer_withID}
  \begin{tabular}{|c|c|c|c|c|c|}

    \toprule
    Quest.No. & 1 strong disagree & 2 disagree & 3 Neutral & 4 agree & 5 strong agree \\
    \midrule
    $(1)$ & 2 & 0 & 4 & 6 & 5 \\
    $(2)$ & 12 & 3 & 2 & 0 & 0 \\
    $(3)$ & 13 & 2 & 1 & 0 & 0 \\
    \bottomrule

  \end{tabular}
\end{table}

\begin{table}[H]
  \centering
  \caption{Results of the Crossformer (without person ID embedding) in the real-time evaluation}
  \label{tab:Crossformer_withoutID}
  \begin{tabular}{|c|c|c|c|c|c|}

    \toprule
    Quest.No. & 1 strong disagree & 2 disagree & 3 Neutral & 4 agree & 5 strong agree \\
    \midrule
    $(1)$ & 5 & 3 & 6 & 2 & 1 \\
    $(2)$ & 11 & 4 & 2 & 0 & 0 \\
    $(3)$ & 15 & 2 & 0 & 0 & 0 \\
    \bottomrule

  \end{tabular}
\end{table}

\begin{table}[H]
  \centering
  \caption{Results of the Crossformer (with person ID embedding) in the real-time evaluation}
  \label{tab:Crossformer_withID}
  \begin{tabular}{|c|c|c|c|c|c|}

    \toprule
    Quest.No. & 1 strong disagree & 2 disagree & 3 Neutral & 4 agree & 5 strong agree \\
    \midrule
    $(1)$ & 4 & 4 & 6 & 2 & 1 \\
    $(2)$ & 10 & 2 & 2 & 2 & 1 \\
    $(3)$ & 11 & 5 & 1 & 0 & 0 \\
    \bottomrule

  \end{tabular}
\end{table}

\begin{table}[H]
  \centering
  \caption{Statistics Results of the Simple Transformer (without person ID embedding) in the real-time evaluation}
  \label{tab:SsimpleTransformer_withoutID}
  \begin{tabular}{|c|c|c|c|c|c|}

    \toprule
    Quest.No. & Mean & Median & Mode & SD \\
    \midrule
    $(1)$ & 3.9 & 4 & 3 & 1.0 \\
    $(2)$ & 3.0 & 3 & 3 & 1.0 \\
    $(3)$ & 3.2 & 3 & 4 & 0.9 \\
    \bottomrule

  \end{tabular}
\end{table}

\begin{table}[H]
  \centering
  \caption{Statistics Results of the Simple Transformer (with person ID embedding) in the real-time evaluation}
  \label{tab:SsimpleTransformer_withID}
  \begin{tabular}{|c|c|c|c|c|c|}

    \toprule
    Quest.No. & Mean & Median & Mode & SD \\
    \midrule
    $(1)$ & 4.0 & 4 & 4 & 0.9 \\
    $(2)$ & 3.2 & 3 & 3 & 0.9 \\
    $(3)$ & 3.9 & 4 & 4 & 0.6 \\
    \bottomrule

  \end{tabular}
\end{table}

\begin{table}[H]
  \centering
  \caption{Statistics Results of the iTransformer (without person ID embedding) in the real-time evaluation}
  \label{tab:SiTransformer_withoutID}
  \begin{tabular}{|c|c|c|c|c|c|}

    \toprule
    Quest.No. & Mean & Median & Mode & SD \\
    \midrule
    $(1)$ & 3.1 & 3 & 3 & 1.5 \\
    $(2)$ & 1.0 & 1 & 1 & 0.0 \\
    $(3)$ & 1.0 & 1 & 1 & 0.0 \\
    \bottomrule

  \end{tabular}
\end{table}

\begin{table}[H]
  \centering
  \caption{Statistics Results of the iTransformer (with person ID embedding) in the real-time evaluation}
  \label{tab:SiTransformer_withID}
  \begin{tabular}{|c|c|c|c|c|c|}

    \toprule
    Quest.No. & Mean & Median & Mode & SD \\
    \midrule
    $(1)$ & 3.7 & 4 & 4 & 1.2 \\
    $(2)$ & 1.4 & 1 & 1 & 0.7 \\
    $(3)$ & 1.4 & 1 & 1 & 0.8 \\
    \bottomrule

  \end{tabular}
\end{table}

\begin{table}[H]
  \centering
  \caption{Statistics Results of the Crossformer (without person ID embedding) in the real-time evaluation}
  \label{tab:SCrossformer_withoutID}
  \begin{tabular}{|c|c|c|c|c|c|}

    \toprule
    Quest.No. & Mean & Median & Mode & SD \\
    \midrule
    $(1)$ & 2.5 & 3 & 3 & 1.2 \\
    $(2)$ & 1.5 & 1 & 1 & 0.7 \\
    $(3)$ & 1.1 & 1 & 1 & 0.3 \\
    \bottomrule

  \end{tabular}
\end{table}

\begin{table}[H]
  \centering
  \caption{Statistics Results of the Crossformer (with person ID embedding) in the real-time evaluation}
  \label{tab:SCrossformer_withID}
  \begin{tabular}{|c|c|c|c|c|c|}

    \toprule
    Quest.No. & Mean & Median & Mode & SD \\
    \midrule
    $(1)$ & 2.5 & 3 & 3 & 1.1 \\
    $(2)$ & 1.9 & 1 & 1 & 1.3 \\
    $(3)$ & 1.4 & 1 & 1 & 0.6 \\
    \bottomrule

  \end{tabular}
\end{table}

\subsection{Results of the Offline Evaluation} 

Tables~\ref{tab:Offline_simpleTransformer_withoutID} to~\ref{tab:Offline_Crossformer_withID} present the questionnaire results for each model in the Offline evaluation. Table~\ref{tab:Offline_simpleTransformer_withoutID} shows the results for the simple Transformer (without person ID embedding), Table~\ref{tab:Offline_simpleTransformer_withID} for the simple Transformer (with person ID embedding), Table~\ref{tab:Offline_iTransformer_withoutID} for iTransformer (without person ID embedding), Table~\ref{tab:Offline_iTransformer_withID} for iTransformer (with person ID embedding), Table~\ref{tab:Offline_Crossformer_withoutID} for Crossformer (without person ID embedding), and Table~\ref{tab:Offline_Crossformer_withID} for Crossformer (with person ID embedding).

In addition, the corresponding statistical measures are presented in Tables~\ref{tab:Offline_SsimpleTransformer_withoutID} to~\ref{tab:Offline_SCrossformer_withID}.

\begin{table}[H]
  \centering
  \caption{Results of the Simple Transformer (without person ID embedding) in the Offline evaluation}
  \label{tab:Offline_simpleTransformer_withoutID}
  \begin{tabular}{|c|c|c|c|c|c|}

    \toprule
    Quest.No. & 1 strong disagree & 2 disagree & 3 Neutral & 4 agree & 5 strong agree \\
    \midrule
    $(1)$ & 0 & 1 & 1 & 5 & 10 \\
    $(2)$ & 0 & 1 & 4 & 9 & 3 \\
    $(3)$ & 0 & 1 & 2 & 10 & 4 \\
    \bottomrule

  \end{tabular}
\end{table}

\begin{table}[H]
  \centering
  \caption{Results of the Simple Transformer (with person ID embedding) in the Offline evaluation}
  \label{tab:Offline_simpleTransformer_withID}
  \begin{tabular}{|c|c|c|c|c|c|}

    \toprule
    Quest.No. & 1 strong disagree & 2 disagree & 3 Neutral & 4 agree & 5 strong agree \\
    \midrule
    $(1)$ & 0 & 0 & 0 & 6 & 11 \\
    $(2)$ & 0 & 0 & 2 & 10 & 5 \\
    $(3)$ & 0 & 0 & 0 & 5 & 12 \\
    \bottomrule

  \end{tabular}
\end{table}

\begin{table}[H]
  \centering
  \caption{Results of the iTransformer (without person ID embedding) in the Offline evaluation}
  \label{tab:Offline_iTransformer_withoutID}
  \begin{tabular}{|c|c|c|c|c|c|}

    \toprule
    Quest.No. & 1 strong disagree & 2 disagree & 3 Neutral & 4 agree & 5 strong agree \\
    \midrule
    $(1)$ & 6 & 1 & 2 & 2 & 6 \\
    $(2)$ & 15 & 1 & 1 & 0 & 0 \\
    $(3)$ & 17 & 0 & 0 & 0 & 0 \\
    \bottomrule

  \end{tabular}
\end{table}

\begin{table}[H]
  \centering
  \caption{Results of the iTransformer (with person ID embedding) in the Offline evaluation}
  \label{tab:Offline_iTransformer_withID}
  \begin{tabular}{|c|c|c|c|c|c|}

    \toprule
    Quest.No. & 1 strong disagree & 2 disagree & 3 Neutral & 4 agree & 5 strong agree \\
    \midrule
    $(1)$ & 2 & 2 & 4 & 3 & 6 \\
    $(2)$ & 11 & 4 & 2 & 0 & 0 \\
    $(3)$ & 14 & 2 & 1 & 0 & 0 \\
    \bottomrule

  \end{tabular}
\end{table}

\begin{table}[H]
  \centering
  \caption{Results of the Crossformer (without person ID embedding) in the Offline evaluation}
  \label{tab:Offline_Crossformer_withoutID}
  \begin{tabular}{|c|c|c|c|c|c|}

    \toprule
    Quest.No. & 1 strong disagree & 2 disagree & 3 Neutral & 4 agree & 5 strong agree \\
    \midrule
    $(1)$ & 1 & 3 & 3 & 4 & 6 \\
    $(2)$ & 11 & 2 & 3 & 1 & 0 \\
    $(3)$ & 14 & 3 & 0 & 0 & 0 \\
    \bottomrule

  \end{tabular}
\end{table}

\begin{table}[H]
  \centering
  \caption{Results of the Crossformer (with person ID embedding) in the Offline evaluation}
  \label{tab:Offline_Crossformer_withID}
  \begin{tabular}{|c|c|c|c|c|c|}

    \toprule
    Quest.No. & 1 strong disagree & 2 disagree & 3 Neutral & 4 agree & 5 strong agree \\
    \midrule
    $(1)$ & 3 & 3 & 3 & 1 & 7 \\
    $(2)$ & 11 & 3 & 0 & 2 & 1 \\
    $(3)$ & 8 & 7 & 2 & 0 & 0 \\
    \bottomrule

  \end{tabular}
\end{table}

\begin{table}[H]
  \centering
  \caption{Statistics Results of the Simple Transformer (without person ID embedding) in the Offline evaluation}
  \label{tab:Offline_SsimpleTransformer_withoutID}
  \begin{tabular}{|c|c|c|c|c|c|}

    \toprule
    Quest.No. & Mean & Median & Mode & SD \\
    \midrule
    $(1)$ & 4.4 & 5 & 5 & 0.8 \\
    $(2)$ & 3.8 & 4 & 4 & 0.8 \\
    $(3)$ & 4.0 & 4 & 4 & 0.8 \\
    \bottomrule

  \end{tabular}
\end{table}

\begin{table}[H]
  \centering
  \caption{Statistics Results of the Simple Transformer (with person ID embedding) in the Offline evaluation}
  \label{tab:Offline_SsimpleTransformer_withID}
  \begin{tabular}{|c|c|c|c|c|c|}

    \toprule
    Quest.No. & Mean & Median & Mode & SD \\
    \midrule
    $(1)$ & 4.6 & 5 & 5 & 0.5 \\
    $(2)$ & 4.2 & 4 & 4 & 0.6 \\
    $(3)$ & 4.7 & 5 & 5 & 0.5 \\
    \bottomrule

  \end{tabular}
\end{table}

\begin{table}[H]
  \centering
  \caption{Statistics Results of the iTransformer (without person ID embedding) in the Offline evaluation}
  \label{tab:Offline_SiTransformer_withoutID}
  \begin{tabular}{|c|c|c|c|c|c|}

    \toprule
    Quest.No. & Mean & Median & Mode & SD \\
    \midrule
    $(1)$ & 3.1 & 3 & 1 & 1.7 \\
    $(2)$ & 1.2 & 1 & 1 & 0.5 \\
    $(3)$ & 1.0 & 1 & 1 & 0.0 \\
    \bottomrule

  \end{tabular}
\end{table}

\begin{table}[H]
  \centering
  \caption{Statistics Results of the iTransformer (with person ID embedding) in the Offline evaluation}
  \label{tab:Offline_SiTransformer_withID}
  \begin{tabular}{|c|c|c|c|c|c|}

    \toprule
    Quest.No. & Mean & Median & Mode & SD \\
    \midrule
    $(1)$ & 3.5 & 4 & 5 & 1.4 \\
    $(2)$ & 1.5 & 1 & 1 & 0.7 \\
    $(3)$ & 1.2 & 1 & 1 & 0.5 \\
    \bottomrule

  \end{tabular}
\end{table}

\begin{table}[H]
  \centering
  \caption{Statistics Results of the Crossformer (without person ID embedding) in the Offline evaluation}
  \label{tab:Offline_SCrossformer_withoutID}
  \begin{tabular}{|c|c|c|c|c|c|}

    \toprule
    Quest.No. & Mean & Median & Mode & SD \\
    \midrule
    $(1)$ & 3.6 & 4 & 5 & 1.3 \\
    $(2)$ & 1.6 & 1 & 1 & 1.0 \\
    $(3)$ & 1.2 & 1 & 1 & 0.4 \\
    \bottomrule

  \end{tabular}
\end{table}

\begin{table}[H]
  \centering
  \caption{Statistics Results of the Crossformer (with person ID embedding) in the Offline evaluation}
  \label{tab:Offline_SCrossformer_withID}
  \begin{tabular}{|c|c|c|c|c|c|}

    \toprule
    Quest.No. & Mean & Median & Mode & SD \\
    \midrule
    $(1)$ & 3.4 & 3 & 5 & 1.6 \\
    $(2)$ & 1.8 & 1 & 1 & 1.3 \\
    $(3)$ & 1.6 & 2 & 1 & 0.7 \\
    \bottomrule

  \end{tabular}
\end{table}

\section{Discussion}
In this section, we discuss the results based on the questionnaire regarding motion generation. Although the questionnaire results were presented in detail in Section 5, Table~\ref{tab:results_summary} summarizes only the average values for each generation pattern to facilitate the discussion in this section.  Furthermore, Figure~\ref{fig:appendix_offline_step100} in Appendix A presents visualizations in Unity that include the counterpart motions generated by each model in the offline evaluation; please refer to it as needed.

\begin{table}[t]
  \centering
  \caption{Average evaluation scores for each generation pattern (summary)}
  \label{tab:results_summary}
  \begin{tabular}{lccc}
    \hline
    Generation Method \hspace{3cm} Question No.& (1) & (2) & (3) \\
    \hline
    Sim.Transformer (w/o ID embed, real-time) & 3.9 & 3.0 & 3.2 \\
    Sim.Transformer (w/ ID embed, real-time) & 4.0 & 3.2 & 3.9 \\
    iTransformer (w/o ID embed, real-time) & 3.1 & 1.0 & 1.0 \\
    iTransformer (w/ ID embed, real-time) & 3.7 & 1.4 & 1.4 \\
    Crossformer (w/o ID embed, real-time) & 2.5 & 1.5 & 1.1 \\
    Crossformer (w/ ID embed, real-time) & 2.5 & 1.9 & 1.4 \\
    Sim.Transformer (w/o ID embed, offline) & 4.4 & 3.8 & 4.0 \\
    Sim.Transformer (w/ ID embed, offline) & 4.6 & 4.2 & 4.7 \\
    iTransformer (w/o ID embed, offline) & 3.1 & 1.2 & 1.0 \\
    iTransformer (w/ ID embed, offline) & 3.5 & 1.5 & 1.2 \\
    Crossformer (w/o ID embed, offline) & 3.6 & 1.6 & 1.2 \\
    Crossformer (w/ ID embed, offline) & 3.4 & 1.8 & 1.6 \\
    \hline
  \end{tabular}
\end{table}

\subsection{Performance Comparison of Each Model}

In this section, we compare and discuss the overall performance of each model.

First, regarding Question (1), ``Are the transitions between frames of the generated counterpart motion temporally smooth?'', relatively high evaluation scores were obtained overall. This result indicates that the joint positions of the generated motion change continuously across frames, suggesting that the models were able to generate motions while referencing past frames. However, compared with the simple Transformer, both iTransformer and Crossformer tend to show slightly lower scores. The reason for this will be discussed later in relation to Question (3).

Before discussing Question (2), we first consider the results of Question (3), which are closely related to its interpretation. For Question (3), ``Does the generated counterpart motion appear human-like overall?'', the simple Transformer achieved relatively high scores. This suggests that the simple Transformer was able to generate motions that maintain plausible body posture. In contrast, iTransformer and Crossformer received overall low scores. This indicates that these models tend to produce motions with collapsed or unstable postures. A similar tendency was observed in Question (1), where these two models also showed lower scores. This can be attributed to the accumulation of errors during motion generation, which gradually degrades posture and results in larger frame-to-frame joint displacements. Furthermore, in Question (3), Crossformer achieved slightly higher scores than iTransformer, suggesting that Crossformer exhibits less severe posture degradation than iTransformer.

Next, regarding Question (2), ``Is the generated counterpart motion a coherent and appropriate response to the subject's motion?'', the simple Transformer tends to achieve higher scores. This suggests that, although it does not always generate perfectly corresponding boxing motions, it is capable of expressing interaction patterns such as producing some form of responsive action when the subject performs a punch. In contrast, iTransformer and Crossformer obtained considerably lower scores overall. This is likely because posture collapse makes it difficult to perceive a clear response to the subject's motion, resulting in lower evaluations. However, Crossformer again shows slightly higher scores than iTransformer, indicating that Crossformer may be able to represent interaction between the subject's motion and the counterpart's motion to some extent.

\subsection{Effect of Person ID Embedding on Evaluation Results}

In this section, we discuss the effect of the presence or absence of person ID embedding on the evaluation results during model training.

Compared to the case without person ID embedding, the use of person ID embedding shows a tendency to increase evaluation scores for almost all generation patterns across all evaluation criteria. In particular, a notable improvement is observed in Question (3) for the simple Transformer. This is likely because person ID embedding explicitly distinguishes which joints belong to which individual, enabling more stable learning of motion distributions for each person and thereby suppressing posture degradation.

In addition, for Question (2), a slight increase in evaluation scores is observed across all generation patterns. From these results, it can be inferred that person ID embedding does not negatively affect the learning of interactions between different individuals and may provide a certain level of improvement.

\subsection{Effect of Temporal Constraints on Evaluation Results}

In this section, we compare and discuss the evaluation scores of each generation pattern obtained from the real-time evaluation and the offline evaluation.

Compared with the real-time evaluation, the offline evaluation shows a tendency for higher scores across almost all generation patterns for all evaluation criteria. This suggests that, in the real-time evaluation, inference latency of the model and delays caused by the system used for real-time visualization of motion negatively affected the evaluation results.

In Question (1), a particularly large increase is observed for the Crossformer. This is likely because Crossformer requires a longer inference time than the other models; therefore, in the offline setting, this delay is eliminated, resulting in smoother joint movements when the motion is visualized.

For Question (2) and (3), the simple Transformer shows a notable increase in evaluation scores. One possible reason for this is the difference in frame rate between the motion data used for training and the motion data input in real time, which may have affected the accuracy of motion generation.

\subsection{Discussion Based on Model Learning Characteristics}

From the evaluation results, it was found that the simple Transformer does not suffer from posture collapse and is capable of representing a certain degree of interaction, although it does not generate perfectly corresponding boxing motions. In contrast, iTransformer and Crossformer accumulate errors at each frame during motion generation, which eventually leads to posture collapse. In this section, we discuss the causes of this phenomenon from the perspective of the learning characteristics of each model.

A key difference between the learning mechanisms of iTransformer and Crossformer and that of the simple Transformer lies in their input representations. The simple Transformer takes as input time-series vectors whose elements consist of all feature values at each time step. In contrast, iTransformer and Crossformer take as input feature-wise vectors whose elements consist of temporal values for each feature. As a result, although the simple Transformer does not explicitly learn inter-joint relationships through attention, it updates all joints simultaneously without separating them. This implicitly maintains the global consistency of the pose and prevents prediction errors in individual joints from being independently amplified, thereby avoiding posture collapse.

On the other hand, in iTransformer, each feature is treated as an independent token, and each joint tends to be updated independently along the temporal dimension. This weakens the constraints among joints within the same frame, leading to the accumulation of joint-wise errors and ultimately causing posture collapse.

Similarly, Crossformer employs a structure that models temporal and feature-wise dependencies in a staged manner using Two-Stage Attention. However, unlike the simple Transformer, it does not update all features within the same frame simultaneously as a single token. As a result, posture collapse occurs in a manner similar to that observed in iTransformer. This suggests that modeling feature-wise dependencies alone is insufficient for stable interaction-aware motion generation.

The reason why Crossformer exhibits less severe posture collapse than iTransformer is likely due to the staged information propagation enabled by the Two-Stage Attention mechanism. While iTransformer computes fully connected self-attention across the feature dimension, temporal dependencies are largely handled by linear projection, making it prone to error accumulation along the temporal dimension during autoregressive generation. In contrast, Crossformer first captures temporal dependencies through attention along the time dimension (Cross-Time Stage), and then aggregates and distributes inter-feature information through attention across the feature dimension (Cross-Dimension Stage). This allows for a more balanced integration of temporal and inter-joint information, thereby mitigating error accumulation during sequential generation. As a result, Crossformer exhibits less severe posture degradation than iTransformer.

\section{Conclusion}
In this study, we developed and evaluated Transformer-based models for generating a counterpart's motion in response to a subject's motion. We investigated three architectures: a simple Transformer that models temporal dependencies, iTransformer that captures inter-feature dependencies, and Crossformer that jointly models temporal and feature-wise relationships. We also introduced a person ID embedding to explicitly distinguish between individuals and examined its effect on interaction-aware motion generation, along with evaluations under both real-time and offline settings.
Experimental results show that the simple Transformer can generate plausible interaction-aware motions without suffering from posture collapse, although it does not perfectly reproduce boxing-specific reactions. In contrast, iTransformer and Crossformer accumulate errors over time, leading to unstable motion generation and eventual posture collapse. This suggests that modeling feature-wise dependencies alone is insufficient for stable interaction-aware motion generation. Furthermore, the person ID embedding plays a critical role in maintaining the structural consistency of the human body while preserving interaction learning, resulting in improved motion stability and performance.
Finally, we observed that inference latency and system delays in real-time settings negatively affect motion generation accuracy, highlighting the importance of temporal constraints in practical applications. These findings suggest that explicitly modeling individual identity is essential for stable interaction-aware motion generation, and that model simplicity and stability can be more effective than architectural complexity. Future work includes improving interaction accuracy, extending the method to more diverse scenarios, and reducing latency for real-time deployment.



\begin{appendices}
\section*{Appendix}
\section{Generated Motion Scenes at 100 frames after the start of playback in the offline evaluation}

\begin{figure}[p]
\centering

\begin{subfigure}{0.46\linewidth}
  \centering
  \includegraphics[width=\linewidth]{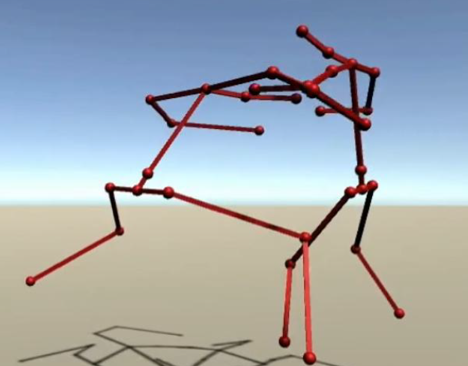}
  \caption{Simple Transformer (w/o ID)}
\end{subfigure}
\hfill
\begin{subfigure}{0.40\linewidth}
  \centering
  \includegraphics[width=\linewidth]{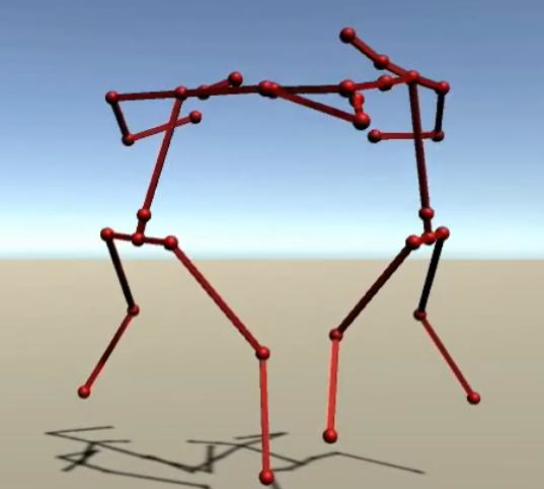}
  \caption{Simple Transformer (w/ ID)}
\end{subfigure}
\hfill

\vspace{5mm}
\begin{subfigure}{0.40\linewidth}
  \centering
  \includegraphics[width=\linewidth]{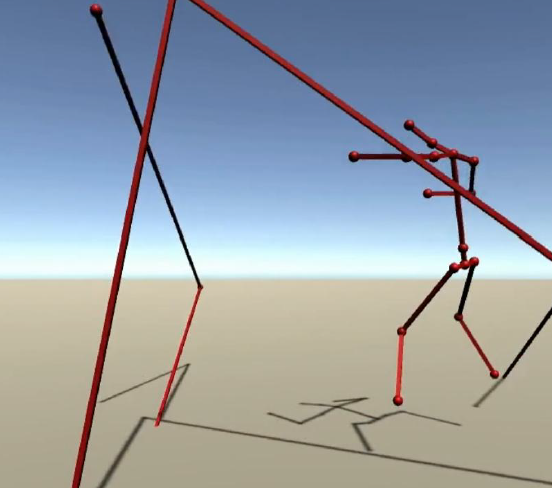}
  \caption{iTransformer (w/o ID)}
\end{subfigure}
\hfill
\begin{subfigure}{0.40\linewidth}
  \centering
  \includegraphics[width=\linewidth]{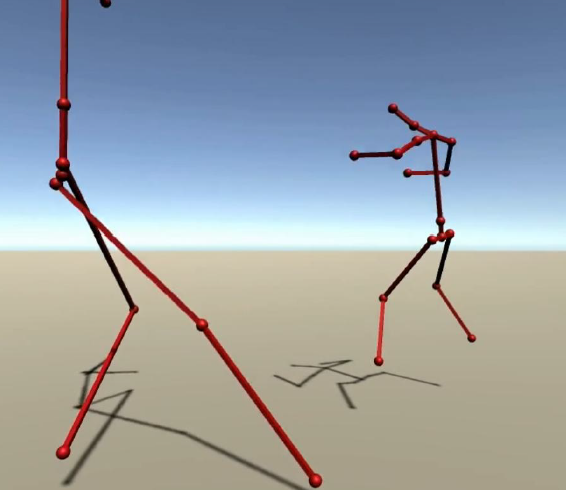}
  \caption{iTransformer (w/ ID)}
\end{subfigure}
\hfill

\vspace{5mm}
\begin{subfigure}{0.40\linewidth}
  \centering
  \includegraphics[width=\linewidth]{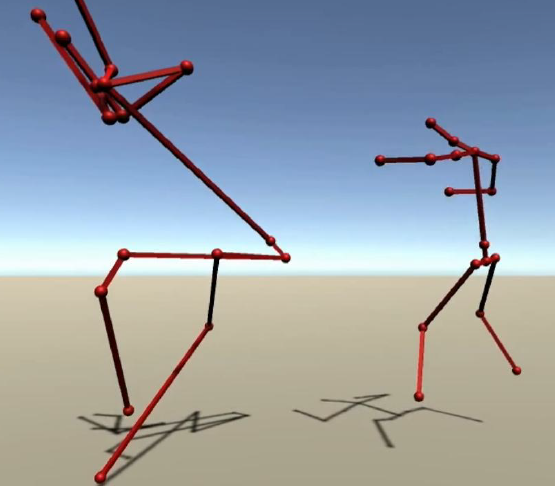}
  \caption{Crossformer (w/o ID)}
\end{subfigure}
\hfill
\begin{subfigure}{0.40\linewidth}
  \centering
  \includegraphics[width=\linewidth]{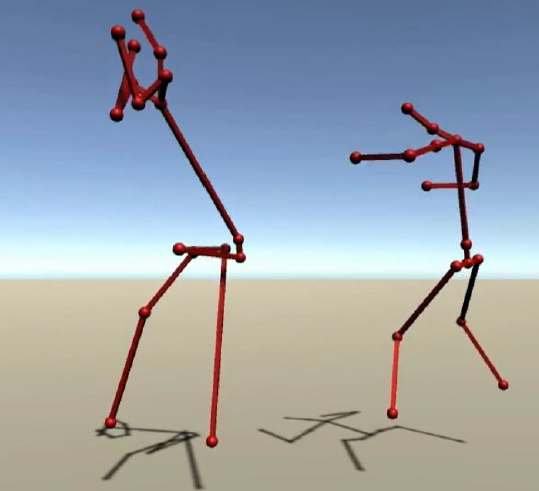}
  \caption{Crossformer (w/ ID)}
\end{subfigure}

\caption{Generated motion scenes at 100 frames after the start of playback in the offline setting. 
Each image shows the subject's motion (right) and the generated counterpart motion (left).}
\label{fig:appendix_offline_step100}

\end{figure}

Figure~\ref{fig:appendix_offline_step100} in this appendix shows generated images in the offline setting. The images are arranged in the same order as lower half of Table~\ref{tab:results_summary}. All images depict scenes at 100 frames after the start of playback. In each scene, the humanoid model on the right represents the subject's motion, whereas the humanoid model on the left represents the generated counterpart motion. In all cases, the model generates the counterpart's motion in response to a straight punch performed by the subject on the right.

First, panels (a) and (b) show results generated by the simple Transformer, where (a) is without person ID embedding and (b) is with person ID embedding. When the subject on the right performs a straight punch, the counterpart on the left avoids the attack by ducking and bending the upper body forward. It can also be seen that, in (b), where person ID embedding is used, the lengths between the leg joints are generated more appropriately than in (a), where person ID embedding is not used.

Next, panels (c) and (d) show results generated by iTransformer, where (c) is without person ID embedding and (d) is with person ID embedding. In both cases, the counterpart on the left exhibits posture collapse due to the accumulation of joint position errors, to the extent that it is no longer clearly recognizable as a human figure. However, the degree of collapse is smaller in (d), with person ID embedding, than in (c), without person ID embedding.

Finally, panels (e) and (f) show results generated by Crossformer, where (e) is without person ID embedding and (f) is with person ID embedding. In both cases, the posture of the counterpart on the left is close to collapsing because of accumulated joint position errors, but the degree of collapse is smaller than that observed in iTransformer. In addition, the degree of collapse is smaller in (f), with person ID embedding, than in (e), without person ID embedding.

\end{appendices}

\end{document}